\documentclass[10pt,twocolumn,letterpaper]{article}

\usepackage[final]{cvpr}      % To produce the REVIEW version
\usepackage{graphicx}
\usepackage{amsmath}
\usepackage{amssymb}
\usepackage{booktabs}

\usepackage[pagebackref,breaklinks,colorlinks]{hyperref}

\usepackage[capitalize]{cleveref}
\crefname{section}{Sec.}{Secs.}
\Crefname{section}{Section}{Sections}
\Crefname{table}{Table}{Tables}
\crefname{table}{Tab.}{Tabs.}

\def\cvprPaperID{11721} % *** Enter the CVPR Paper ID here
\def\confName{CVPR}
\def\confYear{2023}

\begin{document}

\title{ACE: Zero-Shot Image to Image Translation via Pretrained Auto-Contrastive-Encoder}

\author{Sihan XU$^{*}$\\
University of Michigan\\
Ann Arbor
\and
Zelong Jiang$^{*}$\\
University of Michigan\\
Ann Arbor
\and
Ruisi Liu$^{*}$\\
University of Illinois\\
Urbana-Champaign
\and
Kaikai Yang\\
YanShan University\\
Qinhuangdao, China
\and
Zhijie Huang\\
ShanghaiTech University\\
Shanghai, China
}

\maketitle

%%%%%%%%% ABSTRACT
\begin{abstract}
   % The ABSTRACT is to be in fully justified italicized text, at the top of the left-hand column, below the author and affiliation information.
   % Use the word ``Abstract'' as the title, in 12-point Times, boldface type, centered relative to the column, initially capitalized.
   % The abstract is to be in 10-point, single-spaced type.
   % Leave two blank lines after the Abstract, then begin the main text.
   % Look at previous CVPR abstracts to get a feel for style and length.
Image-to-image translation is a fundamental task in computer vision. It transforms images from one domain to images in another domain so that they have particular domain-specific characteristics. 
Most prior works train a generative model to learn the mapping from a source domain to a target domain. However, learning such mapping between domains is challenging because data from different domains can be highly unbalanced in terms of both quality and quantity.
To address this problem, we propose a new approach to extract image features by learning the similarities and differences of samples within the same data distribution via a novel contrastive learning framework, which we call Auto-Contrastive-Encoder (ACE). ACE learns the content code as the similarity between samples with the same content information and different style perturbations. The design of ACE enables us to achieve zero-shot image-to-image translation with no training on image translation tasks for the first time.

Moreover, our learning method can learn the style features of images on different domains effectively. Consequently, our model achieves competitive results on multimodal image translation tasks with zero-shot learning as well. Additionally, we demonstrate the potential of our method in transfer learning. With fine-tuning, the quality of translated images improves in unseen domains. Even though we use contrastive learning, all of our training can be performed on a single GPU with the batch size of 8. Our code is available at \href{https://github.com/SihanXU/ACE}{github.com/SihanXU/ACE}.
\end{abstract}

%%%%%%%%% BODY TEXT
\section{Introduction}
\label{sec:intro}

\begin{figure}
     \centering
     \begin{subfigure}[b]{0.46\textwidth}
         \centering
         \includegraphics[width=\textwidth]{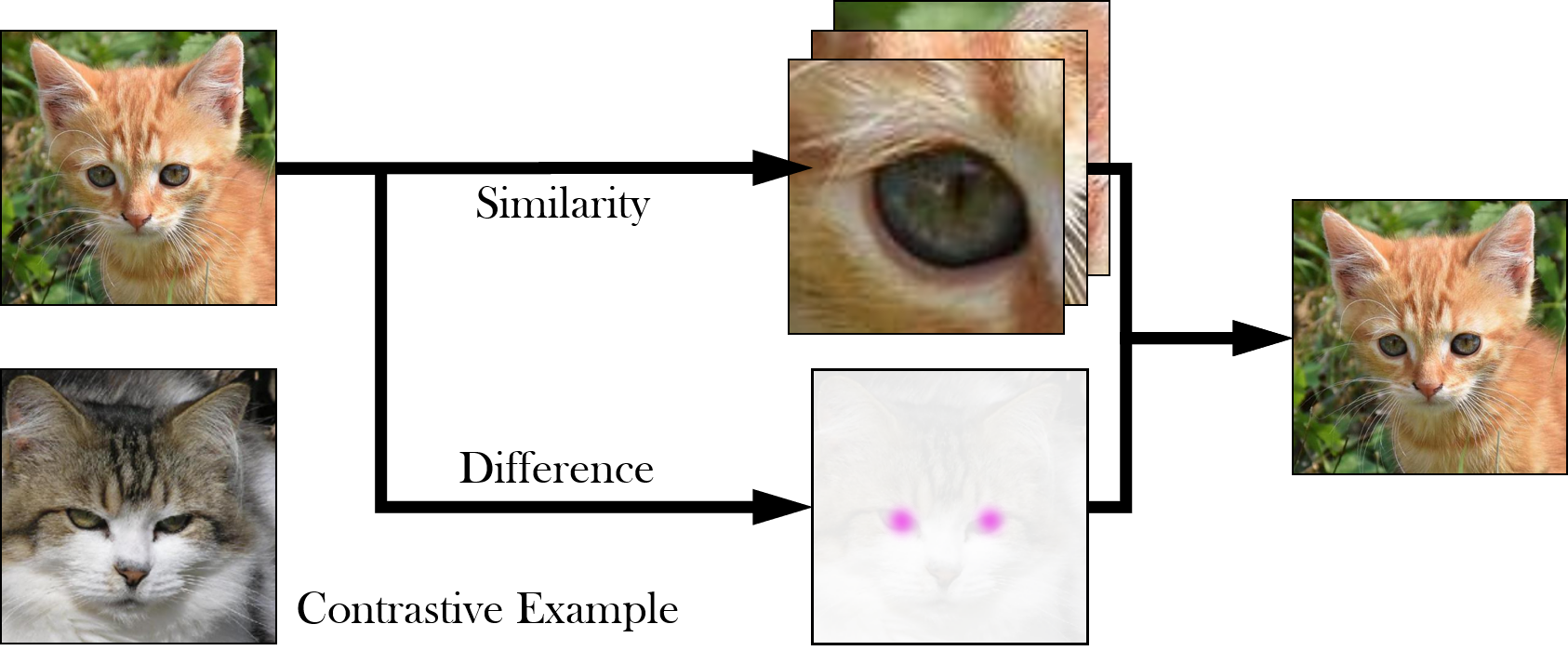}
         \caption{training}
         \label{fig:fig1:training}
         \hspace{20mm}
     \end{subfigure}
     \begin{subfigure}[b]{0.46\textwidth}
         \centering
         \includegraphics[width=\textwidth]{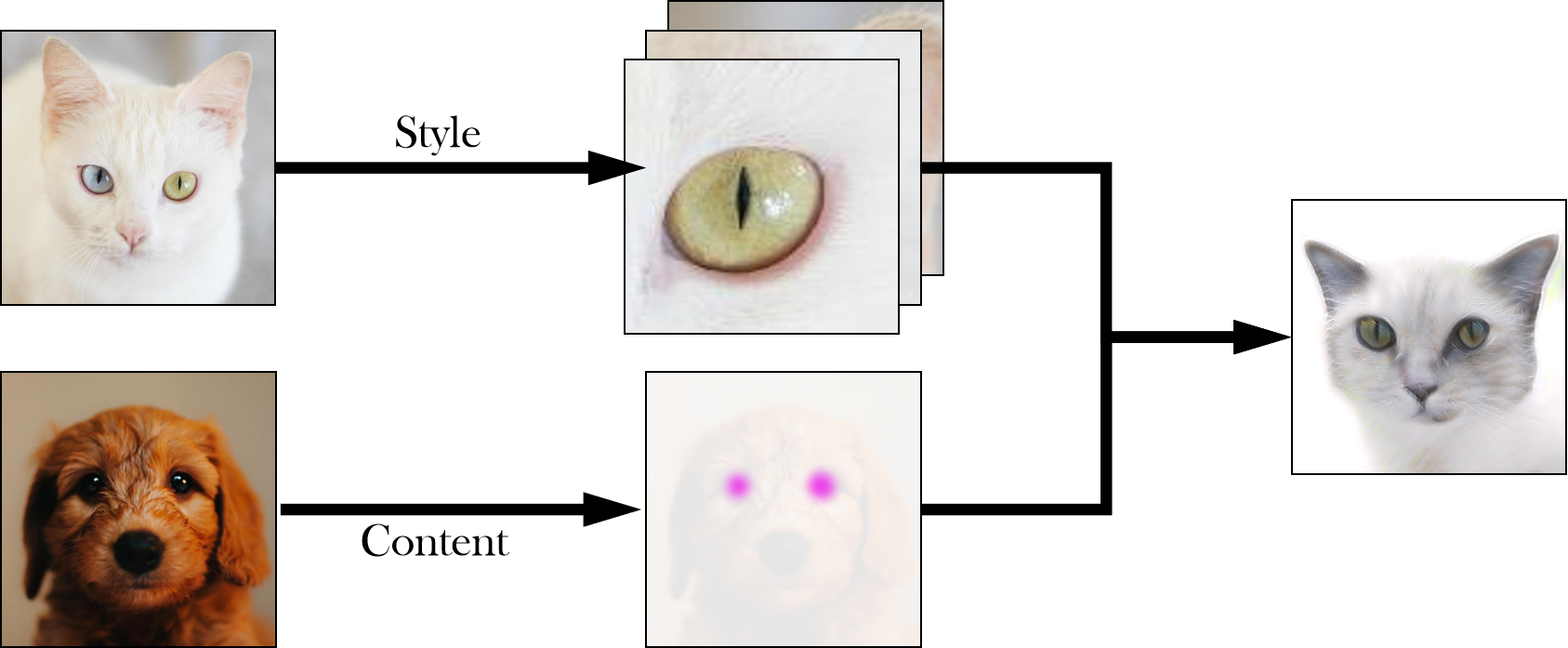}
         \caption{inference}
         \label{fig:fig1:inference}
     \end{subfigure}
        \caption{\textbf{The Main Idea of Auto-Contrastive-Encoder.} When training the ACE model, the encoder learns how to get the similarity and difference from the same domain by contrastive learning. And the decoder uses the similarity features as style and the difference as content to reconstruct the image in both training and inference stage.}
        \label{fig:fig1}
\end{figure}

% \begin{figure}[] %H为当前位置，!htb为忽略美学标准，htbp为浮动图形
% \centering %图片居中
% \includegraphics[width=0.5\textwidth]{Figures/头图.png} %插入图片，[]中设置图片大小，{}中是图片文件名
% \caption{\textbf{The Main Idea of Auto-Contrastive-Encoder.} When training the ACE model, the encoder learns how to get the similarity and difference from the same domain by contrastive learning. And the decoder use the similarity features as style and the difference as content to reconstruct the image in both training and inference stage. 
% } %最终文档中希望显示的图片标题
% \label{fig:fig1} %用于文内引用的标签
% \end{figure}

\begin{figure*}
  \centering % ********
  \begin{subfigure}{0.23\linewidth}
    \includegraphics[width=\textwidth]{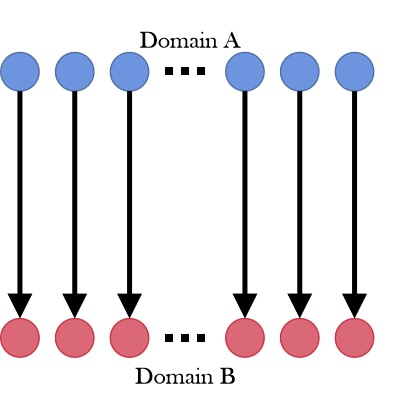}
    \caption{supervised}
    \label{fig:I2IclassA}
  \end{subfigure}
  \hfill %**************
  \begin{subfigure}{0.25\linewidth}
    \includegraphics[width=\textwidth]{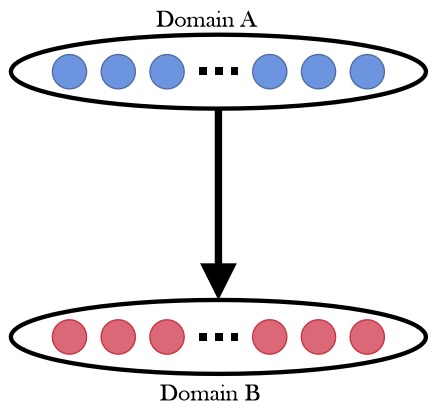}
    \caption{unsupervised}
    \label{fig:I2IclassB}
  \end{subfigure}
  \hfill %**************
  \begin{subfigure}{0.2\linewidth}
    \includegraphics[width=\textwidth]{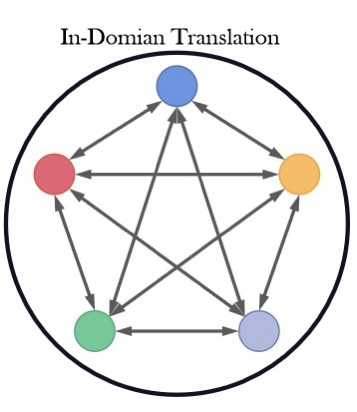}
    \caption{few-shot/truly unsupervised}
    \label{fig:I2IclassC}
  \end{subfigure}
  \hfill %**************
  \begin{subfigure}{0.25\linewidth}
    \includegraphics[width=\textwidth]{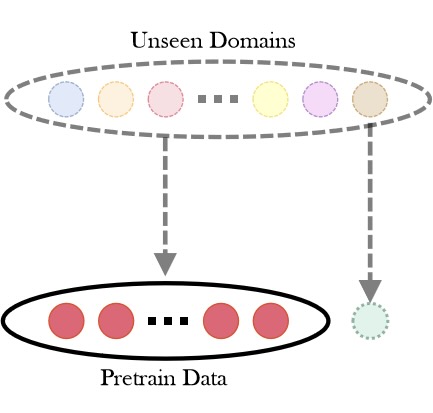}
    \caption{zero-shot (ours)}
    \label{fig:I2IclassD}
  \end{subfigure}
  \caption{\textbf{Different kinds of image to image translation.} (a) Supervised image to image translation like pix2pix \cite{pix2pix} requires paired datas and (b) unsupervised image to image translation like cycleGAN \cite{cycleGAN} requires paired dataset for training. (c) Few-shot \cite{FUNIT} or Truly unsupervised \cite{TUNIT} image to image translation can only translate images in the training domain. Our method, (d) zero-shot image translation is capable to translate images from unseen domains to both of training domain and unseen domains.}
  \label{fig:fig2}
\end{figure*}

% Please follow the steps outlined below when submitting your manuscript to the IEEE Computer Society Press.
% This style guide now has several important modifications (for example, you are no longer warned against the use of sticky tape to attach your artwork to the paper), so all authors should read this new version.
In the field of computer vision, image-to-image translation has been well-established and achieved promising results on various related tasks such as image colorization, style transfer. These existing works are usually achieved by learning the mapping between the source and target domain \cite{pix2pix, cycleGAN, UNIT, MUNIT}. Some new training methods have emerged afterwards \cite{StarGAN, CUT, OST, FUNIT, TUNIT, SwappingAutoencoder}, but they are still trained for learning the mapping relationship, which keeps them from focusing on the distribution of samples. However, such works are inherently fastidious in data distributions. For example, 
pix2pix requires the data of two domains to come in pairs \cite{pix2pix}, while cycleGAN-like methods rely on the joint distribution of two domains\cite{cycleGAN, UNIT, MUNIT}. Some new methods, such as \cite{FUNIT, StarGAN, TUNIT}, still have strict requirements for data. Meanwhile, one-shot image translation (OST) \cite{OST} has only achieved limited breakthroughs with the same idea of learning the mapping. On the other hand, most other works have focused more on how to improve the quality of generated images \cite{CUT, SwappingAutoencoder}. As a consequence, they still fail to move the eyes off the mapping relations to other approaches. Conversely, we believe that we can solve the problem of strict data requirements if there is a training method that can achieve the image translation task without learning the mapping between different distributions. 

In this paper, we propose a novel learning task (\Cref{sec:method}) that implements image translation by learning similar and different features of images within a data distribution (Fig. \ref{fig:fig1}). We note that the features that are similar before and after image translation are precisely the features that need to be retained, while the features that are different under the same distribution are the features that need to be translated. In this way, our model can recognize the features to be retained or transformed by learning the similarities and differences within the distribution. Moreover, such a training method without learning the mapping relation can also translate the samples in unseen domains without being trained on image translation tasks, thereby achieving zero-shot learning as Fig. \ref{fig:fig1:inference}, Fig. \ref{fig:I2IclassD} . Despite several previous works claiming to attain zero-shot image translation \cite{ZUNIT, ZstGAN}, they merely perform style transformation within the features in the specific domain. 

Based on these ideas, we propose the Auto-Contrastive-Encoder (ACE), an Auto-Encoder structure that incorporates contrastive learning (\Cref{sec:method}). Contrastive learning provides us with the effectiveness of learning similarities between positive samples and differences between negative samples, which encourages our model to learn the similar and different features in the same distribution. In this paper, we use a structure similar to Simple Siamese Representation Learning (SimSiam) \cite{SimSiam} to keep the model simple while allowing the model to be trained with a small batch size (\Cref{sec:exp}). For contrastive learning, we propose Adaptive Instance Augmentation and perform contrastive learning directly on the encoded image features. Proven by experiments, our method is able to capture the similarity and difference in image features effectively. 

It is worth noting that ACE is a framework applicable to any model instead of a specific model. Our model in this article uses the VGG \cite{VGG} model for the encoder, while our decoder is a convolutional network (CNN) \cite{CNN} using ResNet \cite{ResNet}. We perform zero-shot image translation and achieve satisfactory results on the Summer$\Leftrightarrow$Winter \cite{MUNIT}, Orange$\Leftrightarrow$Apple \cite{imageNet} and Animal Face \cite{FUNIT} datasets. Furthermore, our experiment on Animal Face \cite{FUNIT} shows that our method also has the potential for transfer learning by pre-training on large datasets. The experiments in this paper can all be completed on a single GPU with batchsize of 8. 

\section{Related Work}

\textbf{Image-to-image translation}. Image-to-image translation, a prevalent problem for computer vision, aims to convert an input image into another output image. It has been applied to style transfer \cite{8732370, styletrans2, gatys2016image}, image denoising \cite{NIPS2012_6cdd60ea, tian2020deep, fan2019brief}, and colorization \cite{Style2Paint, color2, luan2007natural}. Many methods for image translation tasks have been proposed since \cite{pix2pix}. However, these methods are largely limited to learning the mapping between images and rely on the pairs of data in datasets for training \cite{cycleGAN, UNIT, MUNIT, StarGAN, FUNIT, TUNIT, CUT}, and thus can never realize zero-shot learning. Although there is similar research \cite{ZUNIT, ZstGAN} about zero-shot learning methods before, they only do translations in the same domain rather than implement real image translation. 

This paper presents a new method of image translation by learning the similarities and differences of samples within the same distribution. Our ACE approach demonstrates the feasibility of this idea. Our method is not only able to achieve competitive results on zero-shot image-to-image translation but also applicable to various image translation tasks like multimodal translation task. Furthermore, our model has the potential for transfer learning to improve the quality of image translation tasks by pretrain and fine-tune.

\textbf{Contrastive learning} Contrastive learning is an efficient method for unsupervised learning. Its key idea is to learn similar features between positive samples and different features between negative samples\cite{InstDisc}. Based on this idea, several subsequent works with great influence have come into being, such as \cite{Moco, SimCLR, BYOL, SimSiam, Dino}. There are some methods like \cite{BYOL, SimSiam, Dino} can learn similar features between positive samples even without negative samples. 

We use a similar structure to SimSiam \cite{SimSiam} in this paper, but the difference is that we augment the features of images instead of augmenting the image itself. We first use adaptive instance augmentation to augment the features of the images and implement contrastive learning subsequently. Unlike SimSiam \cite{SimSiam}, we also include a predictor in the process of using the encoder. Experiments demonstrate that our method is effective in learning similarities and difference within a distribution. 

\section{Method}
\label{sec:method}
\subsection{Assumption}

\begin{figure}
     \centering
     \begin{subfigure}[b]{0.23\textwidth}
         \centering
         \includegraphics[width=\textwidth]{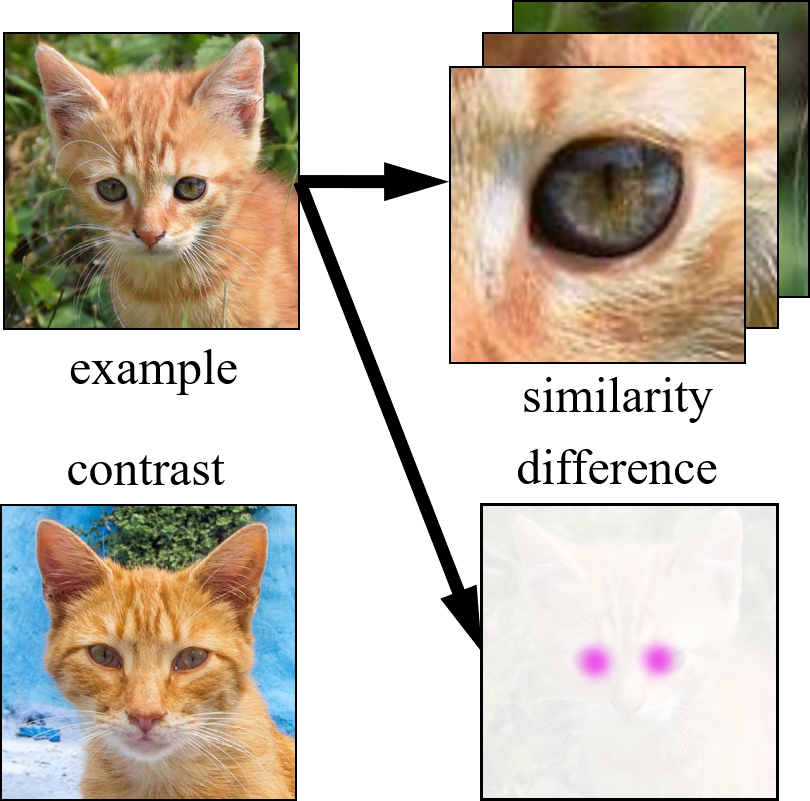}
         \caption{In-domain contrast}
         \label{fig:fig4:in_domain}
     \end{subfigure}
     \begin{subfigure}[b]{0.23\textwidth}
         \centering
         \includegraphics[width=\textwidth]{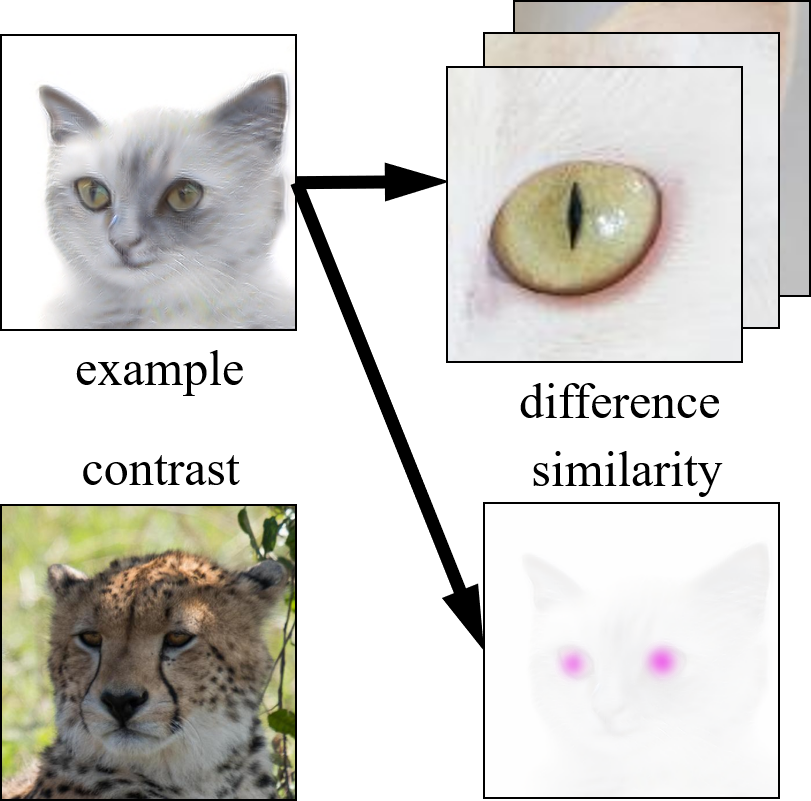}
         \caption{Cross-domain contrast}
         \label{fig:fig4:cross_domain}
     \end{subfigure}
        \caption{\textbf{Assumption of zero-shot translation.} (a) For the examples in the same domain, they should have same style but different content. The similarity should be style and the difference should be the content. (b) For the paired examples before and after translated, they should have same content with different style. Thus, the difference is their style and the similarity is their content.}
        \label{fig:fig4}
\end{figure}

\begin{figure}[] %H为当前位置，!htb为忽略美学标准，htbp为浮动图形
\centering %图片居中
\includegraphics[width=0.45\textwidth]{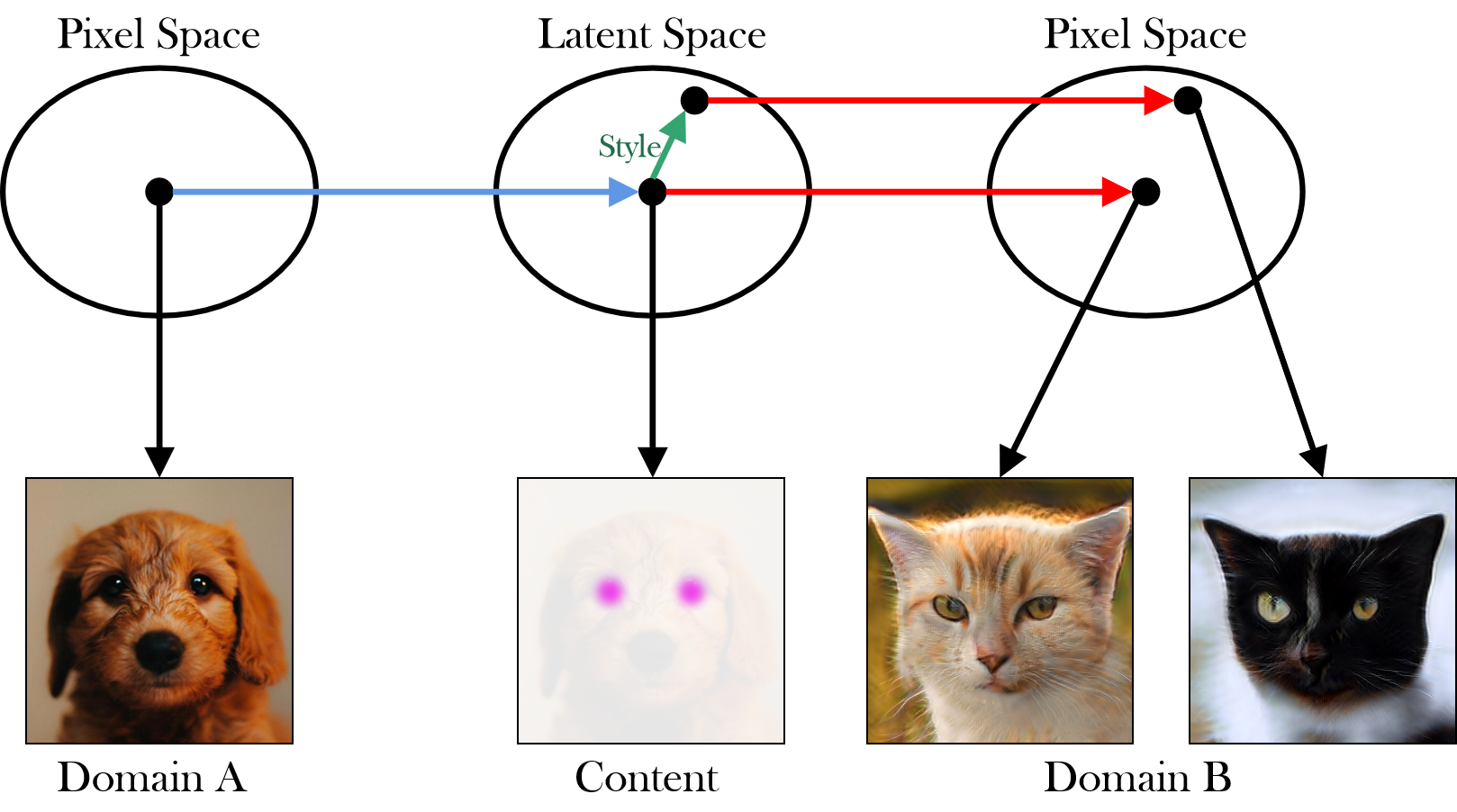} %插入图片，[]中设置图片大小，{}中是图片文件名
\caption{\textbf{Image translation task}. In order to realize the image translation, content code need to be obtained from the pixel space of domain A. And then the output, which is in the pixel space of domain B, should be generated from the content. Additionally, style code is required for multimodal image translation tasks.
} %最终文档中希望显示的图片标题
\label{fig:fig3} %用于文内引用的标签
\end{figure}

Fig. \ref{fig:fig4} shows the fundamental assumption of our method. Based on the effect of in-domain contrast and cross-domain contrast, we can model the content information as the in-domain difference and cross-domain similarity, and the style information as the in-domain similarity and cross-domain difference. Then, we can follow the assumption in MUNIT \cite{MUNIT} like Fig. \ref{fig:fig3} that each image $x_i\in \mathcal X_i$ is generated from a content latent code $c \in \mathcal C$ that is shared by both domains, and a style latent code $s_i \in S_i$ that is specific to the individual domain. For each image, our objective is to find a pair of underlying encoders $E_s$ and $E_c$ to disentangle the two latent codes and a generator $G$ to reconstruct images with these two types of codes. Suppose we have a pair of image $(x_1, x_2)$. We are able to generate a translated image $x_{1\to 2}$ by applying the encoders and the generator, namely $x_{1\to2} = G\left(E_c(x_1), E_s(x_2)\right)$. Note that now $x_{1\to2}$ is also a sample in domain $\mathcal X_2$ with the same content as $x_1$. Then, since the content encoder generates domain-invariant representations, ideally we have $E_c(x_1) = E_c(x_{1\to 2})$. Similarly, considering $x_{1\to 2}$ and $x_2$ are from the same domain, the constraint $E_s(x_2) = E_s(x_{1\to 2})$ should also hold.

\subsection{Model} 

\begin{figure*}
     \centering
     
     \begin{subfigure}[b]{0.6\textwidth}
     \begin{subfigure}[t]{\textwidth}
         \centering
         \includegraphics[width=\textwidth]{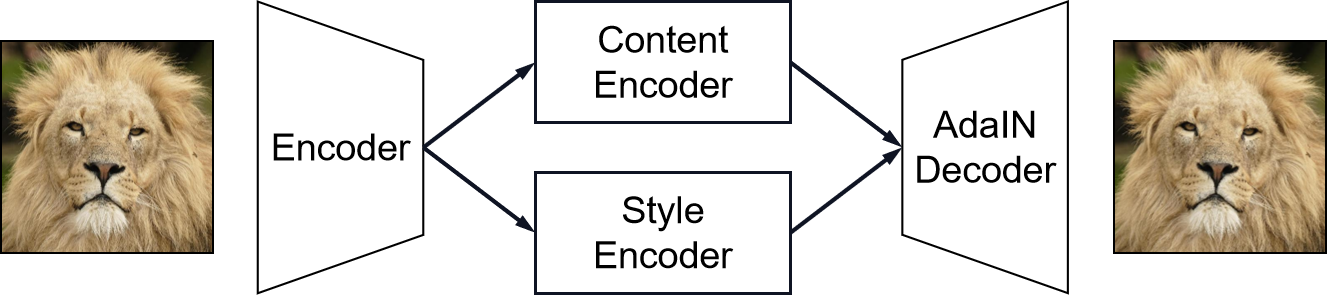}
         \caption{training}
         \label{fig:fig5:train}
     \end{subfigure}\\
     \begin{subfigure}[b]{\textwidth}
         \centering
         \includegraphics[width=\textwidth]{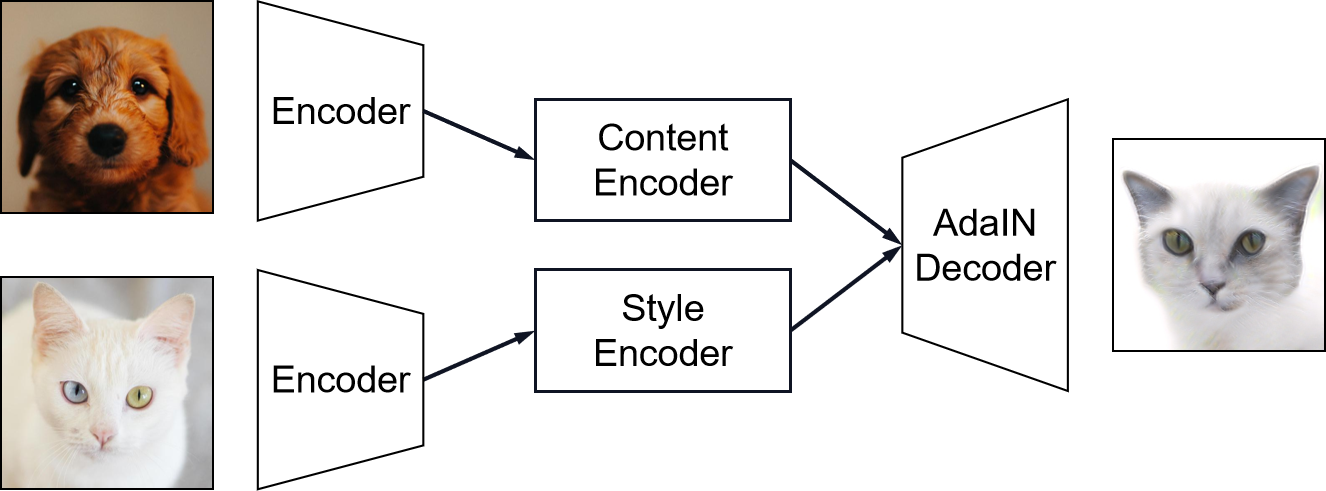}
         \caption{inference}
         \label{fig:fig5:inference}
     \end{subfigure}\hfill
     \end{subfigure}
     \begin{subfigure}[b]{0.35\textwidth}
         \centering
         \includegraphics[width=\textwidth]{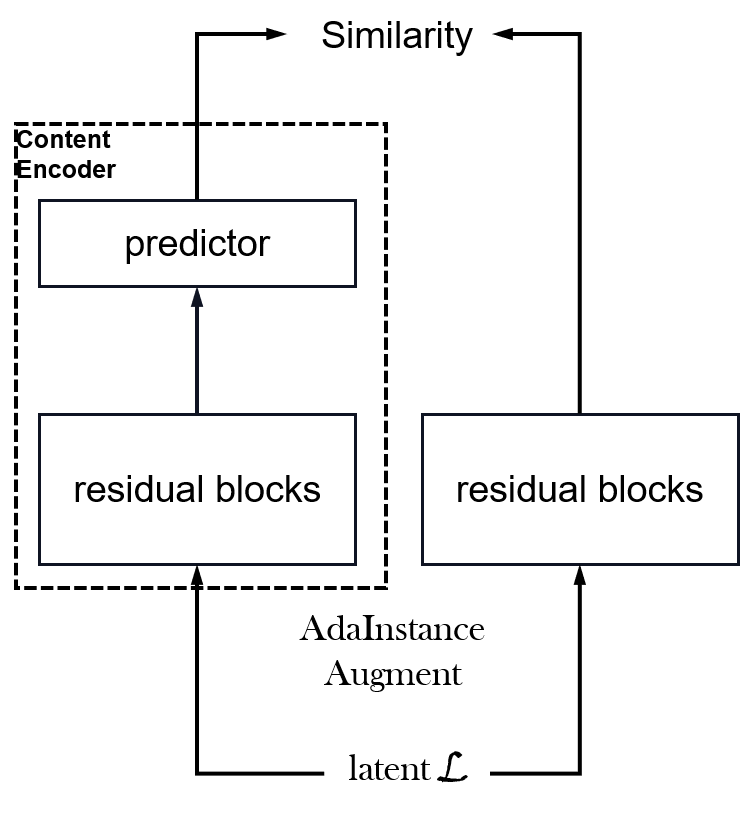}
         \caption{contrastive learning}
         \label{fig:fig5:contrastive}
     \end{subfigure}
        \caption{\textbf{Model overview}. In training, we use (a) auto-encoder and (c) contrastive learning to reconstruct images from the training domain. In inference (b), we use two-stream like architecture to encode content and style, and AdaIn decoder to reconstruct the image. }
        \label{fig:fig5}
\end{figure*}

Fig. \ref{fig:fig5}  is an overview of our model, which is similar to MUNIT \cite{MUNIT}. Our model consists of the encoder, content encoder, style encoder, and decoder. The content encoder contains residual blocks \cite{ResNet} and a predictor, forming a contrastive learning framework. In the process of training (Fig. \ref{fig:fig5:train}), first, we obtain the content and style features respectively through content encoder and style encoder. Finally, the image can be restored by the decoder. In the fine-tuning and inference process (Fig. \ref{fig:fig5:inference}), the encoder obtains features of style and content images so that the style and content images can be learned by the content encoder and style encoder. After these steps, we use the decoder to restore the desired image. 

We use VGG \cite{VGG} as the encoder of the model as in the previous works \cite{AdaIN, MUNIT, FUNIT}. Fig. \ref{fig:fig5:contrastive} presents the learning process of content code. After acquiring the features of input images, we augment the image features with the Adaptive Instance Augmentation. Subsequently, we use a SimSiam similar structure \cite{SimSiam} to implement contrastive learning and obtain the images’ content features. Our style encoder is a single layer CNN with Adaptive Pooling \cite{AdaptivePooling}, which is able to preserve the global feature information of the images. Following MUNIT \cite{MUNIT}, we use an MLP to learn the AdaIN parameters from the style codes.

The content encoder includes two parts as shown in Fig. \ref{fig:fig5:contrastive}. The residual blocks are CNN with skip connection\cite{ResNet} and BatchNorm\cite{BN}. And the predictor is a 2-layer-MLP with a bottleneck, and it has BatchNorm between hidden layers. We don't use the BatchNorm at the output layer.

Our decoder is composed of residual networks and a convolutional network \cite{CNN} with adaptive instance normalization \cite{AdaIN}. As stated in \cite{MUNIT}, instance normalization \cite{IN} and batch normalization would destroy the style features of the image. Therefore, we exclude these two types of normalization in our decoder.

\subsection{Adaptive Instance Augmentation}
 According to the experimental results of using the same adaptive instance normalization (AdaIN) \cite{AdaIN}, the instance norm would affect the style of images. Following this, \cite{MUNIT} proposes to use MLP to dynamically produce the parameters for Instance Normalization layers from style codes. Inspired by these practices, we propose Adaptive Instance Augmentation, where we replace the parameters of AdaIN with Gaussian noises:
\begin{align}
   AdaIN(z, \gamma, \beta) &= \gamma \left(\frac{z - \mu(z)}{\sigma(z)}\right) + \beta,\\ \gamma, \beta &\sim \mathcal N(0,1)
\end{align}
Note that this is an augmentation in the latent space. The procedure to augment a sample $x$ is \begin{equation}
    Aug(x) = G_{\gamma, \beta}(E_c(x)),
\end{equation}
where $G_{\gamma,\beta}$ means that instead of using style encoder, we use the variables $\gamma$ and $\beta$ for the AdaIN layers. We first map $x$ into its content code and then use the randomized AdaIN decoder to reconstruct the augmented sample. This method enables us to modify the style of images while ensuring the same image content. Based on this feature augmentation, our contrastive learning method can make the content encoder insensitive to style features, so that the content feature can be effectively preserved. 

\subsection{Loss Function}
Originating from our assumptions, we first design our loss function to capture the similarity between the content codes with different style features. Let $c_1$ and $c_2$ be two latent codes extracted by content encoder and augmented with Adaptive Instance Augmentation. We define the loss for contrastive learning as the SimSiam loss from \cite{SimSiam}:
\begin{equation}
\begin{aligned}
\mathcal L_{contrast}(c_1, c_2) = \mathcal ContrastiveLoss(p(c_1), stopgrad(c_2))
\end{aligned}
\end{equation}
where $p$ is the predictor layer and $\mathcal ContrastiveLoss$ can be any distance measurement such as negative consine similarity.
The content consistency is also forced by minimizing the distance between content codes extracted from each pair of original sample $x$ and reconstructed sample $x'$:
\begin{equation}
    \mathcal L^{c}_{consist}(x) = \| E_c(x) - E_c(x') \|_1.
\end{equation}
Similarly, for style codes we have 
\begin{equation}
    \mathcal L^{s}_{consist}(x) = \| E_s(x) - E_s(x') \|_1.
\end{equation}

Next, to train the auto-encoder, we adopt a reconstruction loss and a GAN loss to ensure that the reconstructed images follow the distribution of target domain.
\begin{align}
    \mathcal L_{recon}(x) &= \| x - x' \|_1 \\
\mathcal L_{GAN} = \| D(x)& - 1 \|_2^2 + \| D(x') \|_2^2
\end{align}
We train our model with the total objective as the weighted sum of all loss functions mentioned above.

\subsection{Stop gradient}

Since Auto-Encoder updates encoder in the training, it will greatly influence the effect of contrastive learning. Therefore, when we are training the Auto-Encoder, we freeze the content encoder. Which means
\begin{equation}
    x' = G(stopgrad(E_c(x)), E_s(x))
\end{equation}

\subsection{Discriminator}
We use an approach similar to Generative Adversarial Network (GAN) \cite{GAN} to train our ACE to improve the quality of the images. In our experiments, we use a loss function similar to \cite{LSGAN} to make the training more stable; and use SpectralNorm \cite{SpectralNorm} to enable the model to generate images with higher quality.

\section{Experiments}
\label{sec:exp}
\subsection{Implementation Details}

\begin{figure}[] %H为当前位置，!htb为忽略美学标准，htbp为浮动图形
\centering %图片居中
\includegraphics[width=0.3\textwidth]{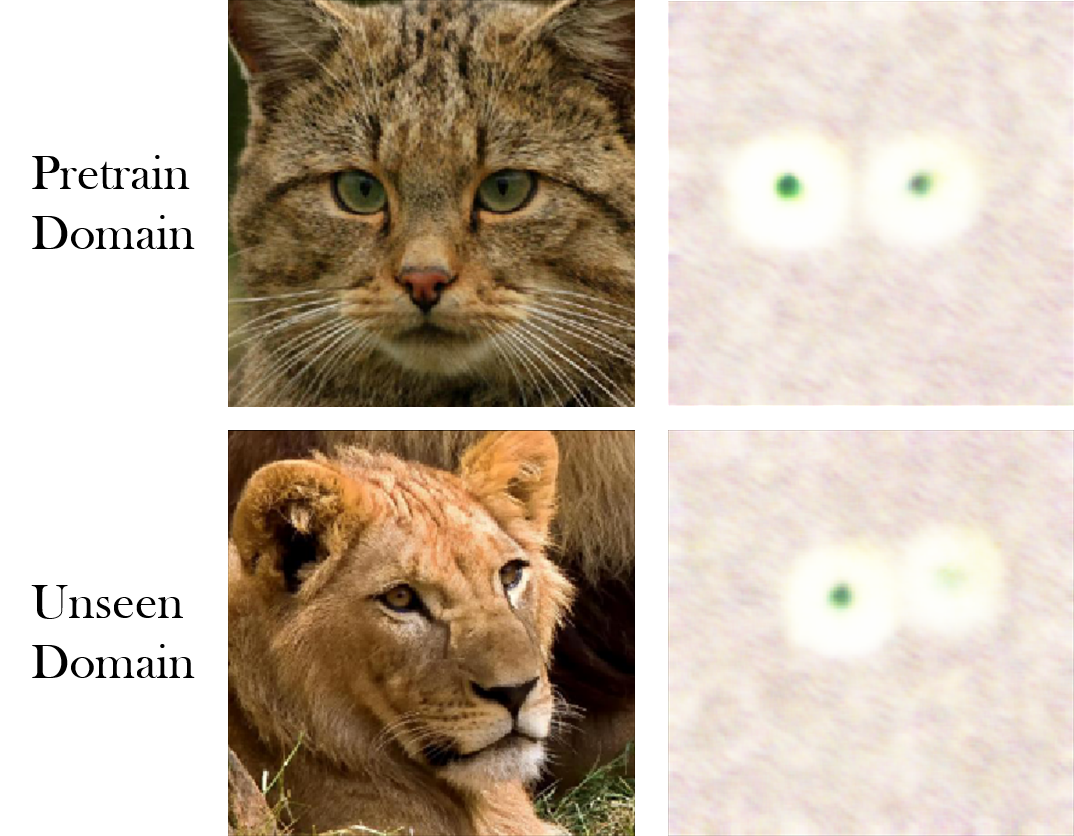} %插入图片，[]中设置图片大小，{}中是图片文件名
\caption{\textbf{Visualization of content code on two domains.} The content code only contains the information of the location of animal's eyes, which is the difference between the in-domain examples. And the content encoder is also capable to encode the content from the unseen domain.} %最终文档中希望显示的图片标题
\label{fig:exp:visualization} %用于文内引用的标签
\end{figure}

Our framework is comprised of a VGG encoder, a content encoder, a style encoder and a decoder. The content encoder consists of four residual blocks and the style encoder contains a global pooling layer and a fully connected layer. For the decoder, we have several residual blocks, each followed by up-sampling layers. We also use Adaptive Instance Normalization layers to dynamically generate parameters of Instance Normalization. However, to accelerate the convergence of style encoder, we propose to use two different style codes to respectively represent the global style in the domain and the individual style of each sample. The domain style code is a learnable tensor which is shared by all data in the pretrain domain, while the individual style code is output by the style encoder. Then we sum these two style codes up before applying them to the AdaIN layers.

\subsection{Datasets}
We conduct the evaluation on the same datasets as \cite{MUNIT,cycleGAN}. Our method achieves satisfying results on Yosemite summer$\Leftrightarrow$winter, apple$\Leftrightarrow$orange and Animal face translation (including data of bigcats, cats and dogs). 

\begin{figure}[] %H为当前位置，!htb为忽略美学标准，htbp为浮动图形
\centering %图片居中
\includegraphics[width=0.45\textwidth]{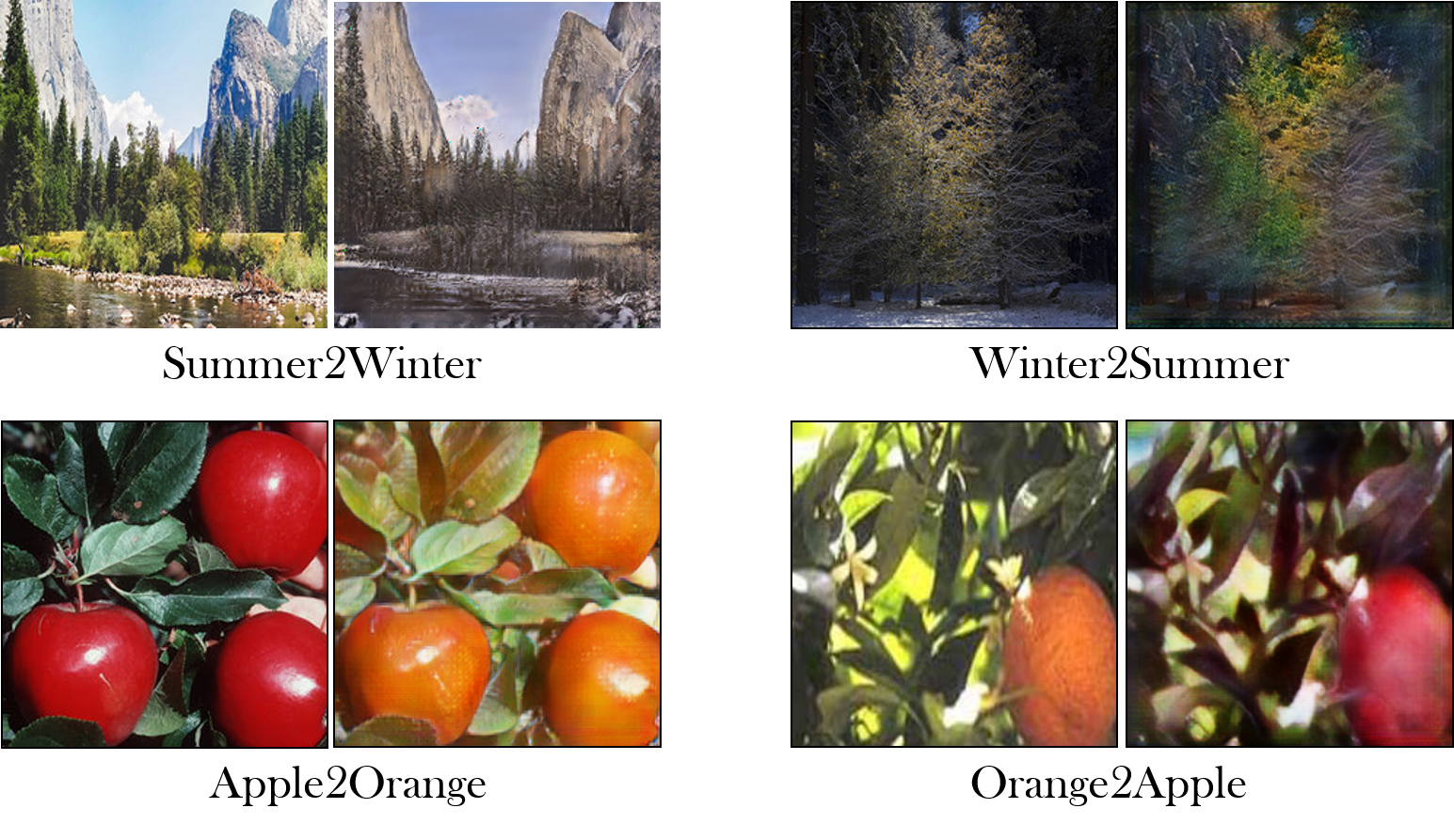} %插入图片，[]中设置图片大小，{}中是图片文件名
\caption{Translation results on Yosemite summer$\Leftrightarrow$winter and Apple$\Leftrightarrow$Orange.} %最终文档中希望显示的图片标题
\label{fig:exp:unimodal} %用于文内引用的标签
\includegraphics[width=0.4\textwidth]{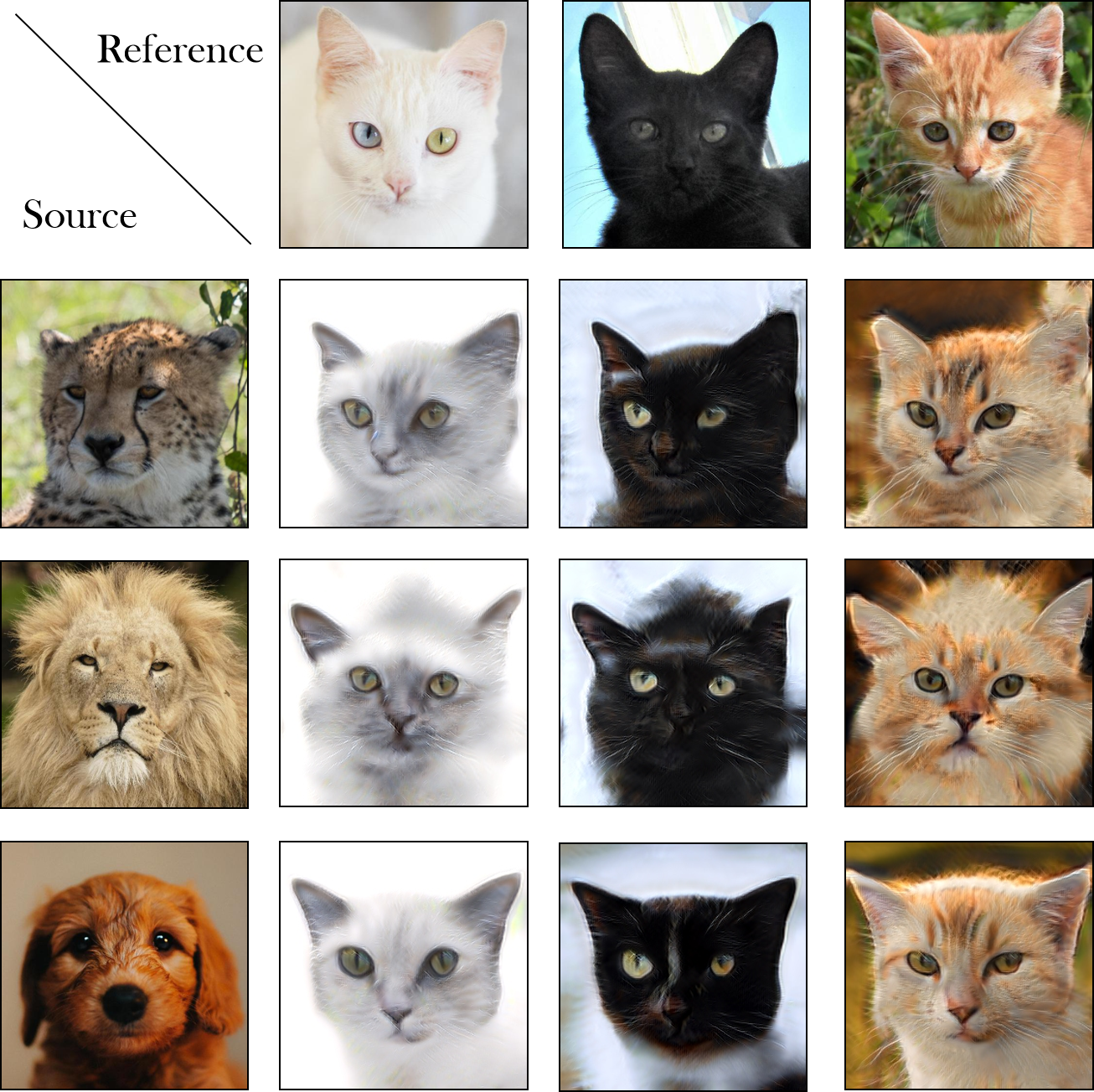} %插入图片，[]中设置图片大小，{}中是图片文件名
\caption{Multimodal Translation results on bigcat$\Leftrightarrow$cat and dog$\Leftrightarrow$cat.} %最终文档中希望显示的图片标题
\label{fig:exp:multimodal} %用于文内引用的标签
\end{figure}

\subsection{Visualization}

\begin{figure*}[] %H为当前位置，!htb为忽略美学标准，htbp为浮动图形
\centering %图片居中
\includegraphics[width=0.8\textwidth]{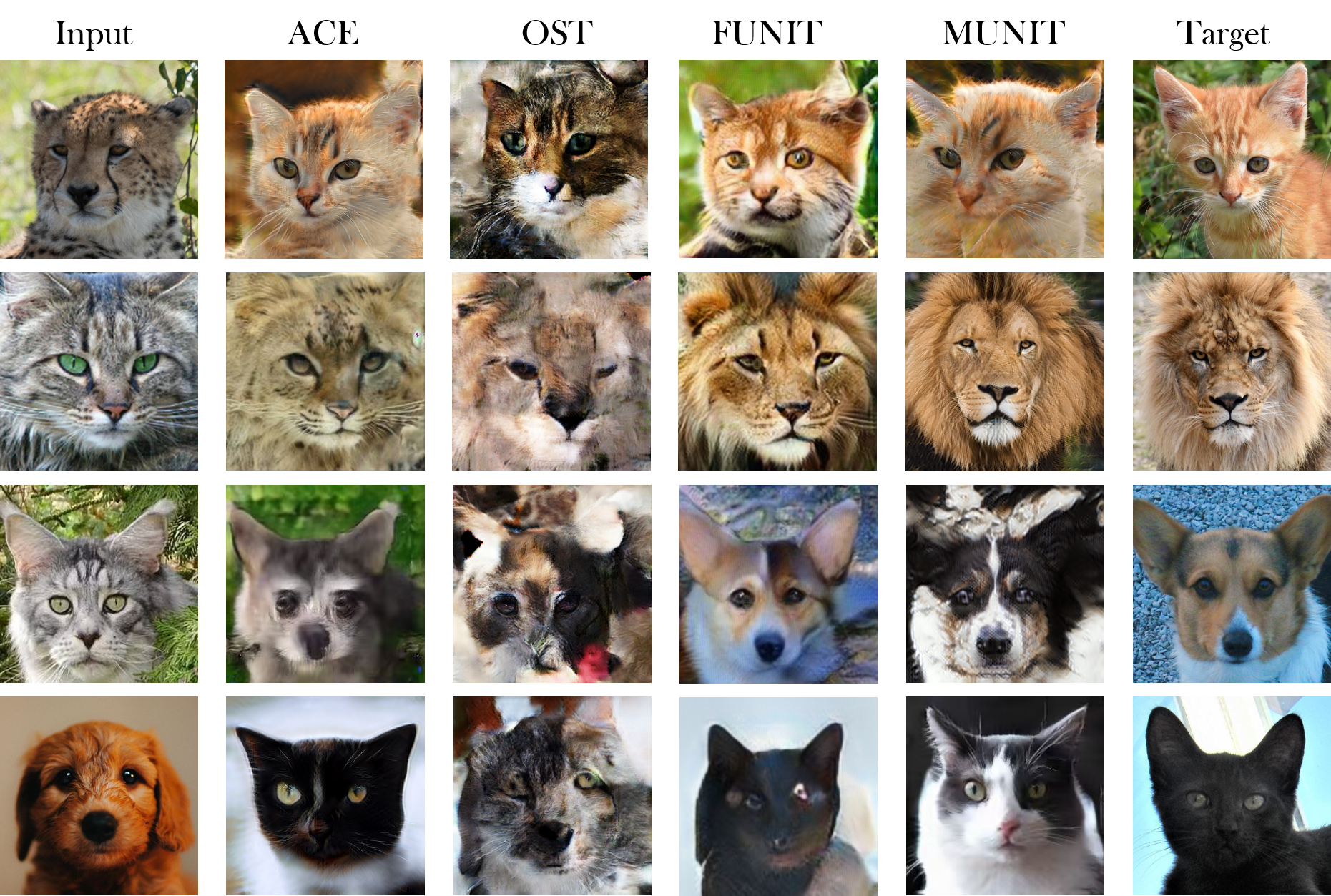} %插入图片，[]中设置图片大小，{}中是图片文件名
\caption{Comparison of zero-shot translation results with one-shot\cite{OST}, few-shot\cite{FUNIT} and unsupervised\cite{MUNIT} methods.} %最终文档中希望显示的图片标题
\label{fig:exp:comparison} %用于文内引用的标签
\end{figure*}

To better understand whether our designed models work as we expect, we adopt some tools of visualization for our extracted content codes. Fig. \ref{fig:exp:visualization} shows the information in the content codes from the pretrain domain and the unseen domain. Both representations indicate the animal's eyes. It verifies that our contrastive learning based encoder is able to extract similar content codes regardless of the domains.

\subsection{Effectiveness of Zero-shot Learning}

Our experiment covers the zero-shot learning on datasets Summer$\Leftrightarrow$Winter \cite{MUNIT} and Orange$\Leftrightarrow$Apple \cite{imageNet}. The final results are shown in Fig. \ref{fig:exp:unimodal}.

At the same time, we conduct multimodal translation on the Animal Image Translation Dataset \cite{FUNIT}. We trained our model on cat dataset and applied to bigcat2cat task and dog2cat task. With the images in Fig. \ref{fig:exp:multimodal}, we can see that our method translates images to different styles while maintaining the original content features.

The focus of our work is not about producing high quality images. For this reason, we only compare our method with OST \cite{OST}, FUNIT \cite{FUNIT}, MUNIT \cite{MUNIT}. During training, our model shares the same settings as MUNIT, while OST and FUNIT keep their settings as mentioned in OST and 
FUNIT. The comparison is presented in Fig. \ref{fig:exp:comparison}. It is apparent that our framework obtains outstanding translating results in a zero-shot manner.

\subsection{Transfer Learning}

\begin{figure}[h] %H为当前位置，!htb为忽略美学标准，htbp为浮动图形
\centering %图片居中
\includegraphics[width=0.43\textwidth]{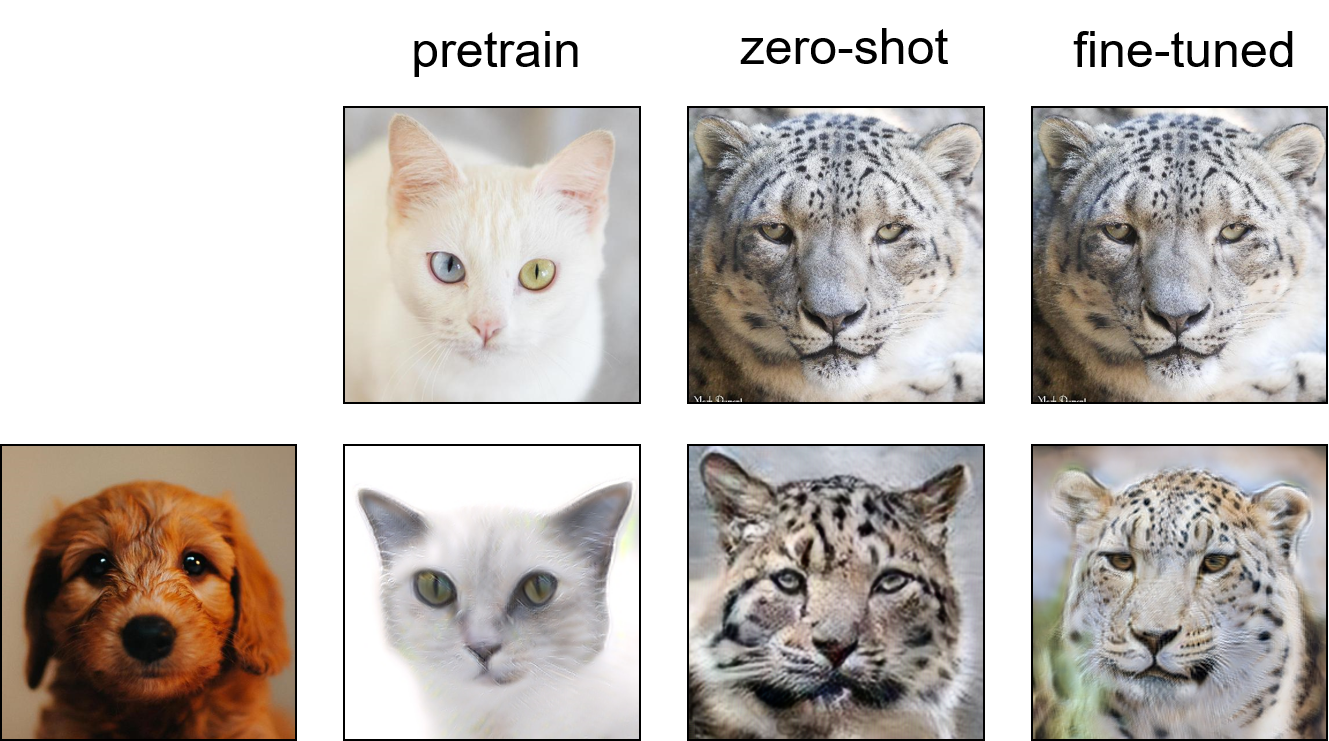} %插入图片，[]中设置图片大小，{}中是图片文件名
\caption{\textbf{Fine-tune result.} Even though ACE is capable of doing the zero-shot translation to an unseen target domain, but the translation result contains some features from the pretrain domain and thus affects the translation image's quality. After fine-tuning on the target domain, ACE can achieve better translation results, which shows the potential of ACE model in transfer learning.} %最终文档中希望显示的图片标题
\label{fig:exp:pretrain} %用于文内引用的标签
\end{figure}

Here we discuss whether our model is suitable for transfer learning. We pre-train our model on a specific domain and test on one or two unseen domains. If we are able to get the training data of the test domains, we can conduct fine-tuning to improve the generated image quality. The process of fine-tuning is quite similar to the pre-training. We use the images in the source domain to generate the content codes and use target domain to get style codes. The fine-tuning loss will be consisted of the latent consistency losses $\mathcal L^c_{consist}, \mathcal L^s_{consist}$ and the GAN loss $\mathcal L_{GAN}$. We don't use the contrastive loss in fine-tuning. 

As shown in Fig. \ref{fig:exp:pretrain}, in experiments of multimodal image translation, we pre-train our model on the cat domain and it achieves satisfactory performance when applied to the task of dog2cat translation. Next, our model still performs well if we change the target domain to the bigcats. Therefore, we believe with pre-training, our model is capable of translating the images from one unseen domain to another unseen domain. If we continue to fine-tune on the dataset of bigcats, we can see that there is much room for improvement of generated image quality. As a result, we believe our model has a great potential for transfer learning based on large datasets like ImageNet\cite{imageNet}.

\subsection{Failure Cases}

\begin{figure}[h] %H为当前位置，!htb为忽略美学标准，htbp为浮动图形
\centering %图片居中
\includegraphics[width=0.45\textwidth]{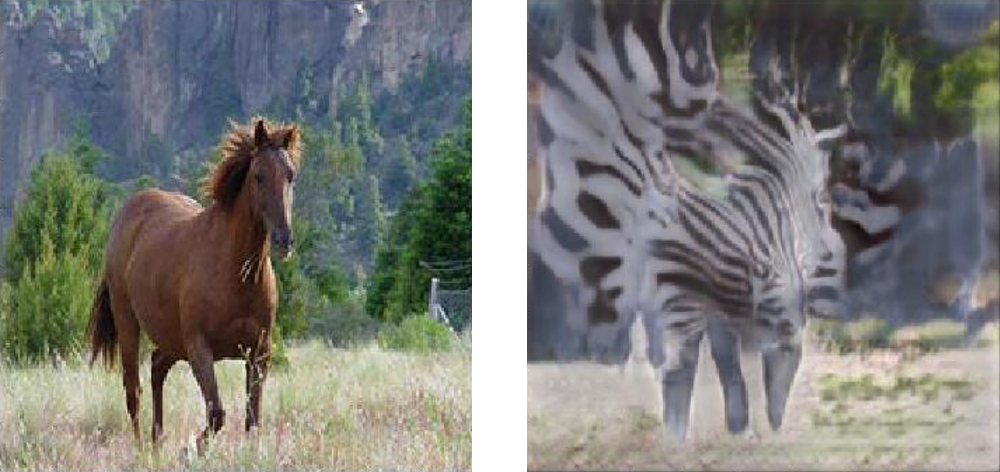} %插入图片，[]中设置图片大小，{}中是图片文件名
\caption{\textbf{Failure Cases.} The context is so complicated that our model cannot integrate style and content information well.} %最终文档中希望显示的图片标题
\label{fig:exp:failure_cases} %用于文内引用的标签
\end{figure}

Our method fails on some cases where the context is complicated such as Fig \ref{fig:exp:failure_cases}. For instance, on the dataset of horse2zebra, our model sometimes erroneously puts the zebra's stripes on the background.

\section{Discussion}

The method proposed in this paper has achieved satisfactory results in the task of image-to-image translation, but we have not conducted experiments on other types of translation tasks. For example, we believe our method can work on language processing as well. Furthermore, due to limited resources, we only tested the potential of our method on transfer learning with small datasets. If we can pre-train on a larger data set, our model may be able to achieve better results in image translation. In this article, we just use a very simple model structure, but our approach is also applicable to other models, such as ResNet \cite{ResNet}, Vision Transformer\cite{ViT} and Diffusion Model \cite{DiffusionModel}. We believe that better results can be achieved if our methods are combined with these further efforts.

\section{Conclusions}

In order to conquer the challenge coming from learning the mapping relationship in image-to-image translation, we propose a new objective, which is to translate images by learning the similarities and differences without learning any mappings or joint distributions. Additionally, we propose a simple model structure called Auto-Contrastive-Encoder to solve this problem, and it has achieved satisfactory results. We have also shown the potential of our model in transfer learning. It is promising that our method can make the task of image-to-image translation move forward.

\label{sec:ccl}

%%%%%%%%% REFERENCES
{\small
\bibliographystyle{ieee_fullname}
\bibliography{egbib}
}

\end{document}